\documentclass[manuscript,screen]{acmart}

\AtBeginDocument{%
  \providecommand\BibTeX{{%
    \normalfont B\kern-0.5em{\scshape i\kern-0.25em b}\kern-0.8em\TeX}}}

\setcopyright{none} % for arXiv

\acmJournal{TRETS}
\usepackage{utfsym}

\usepackage{caption}
\usepackage{subcaption}

\usepackage{amsfonts}
\usepackage{booktabs}
\usepackage{siunitx}

\usepackage{xcolor}

\usepackage{mathtools, nccmath}

 \usepackage{multirow}
 \usepackage{xspace}
\newcommand{\pt}{\ensuremath{p_{\mathrm{T}}}\xspace}

\DeclareMathOperator*{\Int}{Int}
\DeclareMathOperator*{\Clip}{Clip}
\DeclareMathOperator*{\Round}{Round}
\DeclareMathOperator*{\quantize}{quantize}
\DeclareMathOperator*{\dequantize}{dequantize}

\begin{document}

%\title{An end-to-end codesign workflow of Hessian-aware quantized neural networks for FPGAs and ASICs}
\title{End-to-end codesign of Hessian-aware quantized neural networks for FPGAs and ASICs}

\author{Javier Campos}
\email{jcampos@fnal.gov}
\orcid{0009-0008-8029-3267}
\author{Jovan Mitrevski}
\email{jmitrevs@fnal.gov}
\orcid{0000-0001-9098-0513}
\author{Nhan Tran}
\email{ntran@fnal.gov}
\orcid{0000-0002-8440-6854}
\affiliation{%
  \institution{Fermi National Accelerator Laboratory}
  \city{Batavia}
  \state{IL}
  \country{USA}
}
\author{Zhen Dong}
\email{zhendong@berkeley.edu}
\orcid{0000-0002-5951-6170}
\author{Amir Gholami}
\email{amirgh@berkeley.edu}
\orcid{0000-0003-1374-3105}
\additionalaffiliation{
\institution{International Computer Science Institute}
}
\author{Michael W. Mahoney}
\email{mmahoney@stat.berkeley.edu}
\affiliation{%
  \institution{University of California Berkeley}
  \city{Berkeley}
  \state{CA}
  \country{USA}
}
\additionalaffiliation{
\institution{International Computer Science Institute and Lawrence Berkeley National Laboratory}
}
\author{Javier Duarte}
\orcid{0000-0002-5076-7096}
\email{jduarte@ucsd.edu}
\affiliation{%
  \institution{University of California San Diego}
  \city{LA Jolla}
  \state{CA}
  \country{USA}
}

\renewcommand{\shortauthors}{J. Campos, et al.}

\begin{abstract}
  We develop an end-to-end workflow for the training and implementation of co-designed neural networks (NNs) for efficient field-programmable gate array (FPGA) and application-specific integrated circuit (ASIC) hardware.
  Our approach leverages Hessian-aware quantization (HAWQ) of NNs, the Quantized Open Neural Network Exchange (QONNX) intermediate representation, and the hls4ml tool flow for transpiling NNs into FPGA and ASIC firmware.
  This makes efficient NN implementations in hardware accessible to nonexperts, in a single open-sourced workflow that can be deployed for real-time machine-learning applications in a wide range of scientific and industrial settings.
  We demonstrate the workflow in a particle physics application involving trigger decisions that must operate at the 40\,MHz collision rate of the CERN Large Hadron Collider (LHC).
  Given the high collision rate, all data processing must be implemented on custom ASIC and FPGA hardware within the strict area and latency requirements.
  Based on these constraints, we implement an optimized mixed-precision NN classifier for high-momentum particle jets in simulated LHC proton-proton collisions.
\end{abstract}




% \received{}

\maketitle

\clearpage
\tableofcontents
\clearpage

\section{Introduction}
\label{sec:intro}

Machine learning (ML) is pervasive in big data processing, and it is becoming increasingly important as data rates continue to rise.
In particular, ML taking place as close to the data source as possible, or \emph{edge ML}, is increasingly important for both scientific and industrial applications, including applications such as data compression, data volume reduction, and feature extraction for real-time decision-making~\cite{10.3389/fdata.2022.787421}.
Integrating ML at the edge, however, is challenging because of area, power, and latency constraints.
This is especially the case for deep learning (DL) and neural network (NN) models.
Deployment of NNs for edge applications requires carefully-optimized protocols for training as well as finely-tuned implementations for inference.
This typically requires efficient computational platforms such as field-programmable gate arrays (FPGAs) and application-specific integrated circuits (ASICs).
Developing a NN algorithm and implementing it in hardware within system and task constraints is a multistep \textit{codesign} process with a large decision space.
Among other things, this space includes options related to \emph{quantization}, or using reduced precision operations.
In this paper, we present a completely open-source, end-to-end workflow accessible to nonexperts for NN quantization and deployment in FPGAs and ASICs.

Quantization-aware training (QAT) has been shown to be very successful in scaling down model sizes for FPGAs~\cite{jianfei1,jianfei2,jianfei3,huang2021codenet,Coelho2020UltraLL,dong2021hao}.
With QAT, large NNs can be quantized to 8 bits and below, with comparable accuracy to the baseline.
Quantized NNs (QNNs) generally have considerably reduced model sizes and latencies.
Hessian-aware quantization (HAWQ)~\cite{hawqv3} is a mixed-precision integer-only quantization framework for PyTorch~\cite{pytorch} with promising applications.
HAWQ is able to quantize the model to very small bit widths by using mixed-precision guided by second-order (Hessian) information.
In this approach, sensitive layers (determined by Hessian information) are kept at higher precision and insensitive layers are kept at lower precision.
FPGAs are a natural use case for this: they can benefit from this approach since mixed-precision computations are much better supported by FPGAs than other hardware such as GPUs.

While these features make HAWQ an interesting choice for QAT with FPGAs, there does not currently exist a streamlined process to deploy it onto FPGAs directly.
To address this, we introduce additional functionality to HAWQ in order to export QNNs as Quantized Open Neural Network Exchange (QONNX)~\cite{Pappalardo:2022nxk} intermediate representations.
Then the QONNX representation can be ingested by hls4ml~\cite{fast-inference}, an open-source Python library for NN translation and deployment in FPGA and ASIC hardware.
The hls4ml package is designed to be accessible for both hardware experts and nonexperts, and it is flexible enough to deploy QNNs with a broad range of quantization bit widths on different FPGA and ASIC platforms.
It is a popular tool for both scientific and industry edge ML applications~\cite{Loncar:2020hqp, Ghielmetti:2022ndm, Aarrestad:2021zos}.

To demonstrate the performance of our end-to-end workflow, we develop a NN for real-time decision-making in particle physics.
The CERN Large Hadron Collider (LHC) is the world's largest and most powerful particle accelerator.
Particles collide in detectors every 25\,ns, producing tens of terabytes of data.
Because of storage capacity and processing limitations, not every collision event can be recorded.
In these experiments, the online trigger system filters data and stores only the most ``interesting'' events for offline analysis.
Typically, the trigger system uses simple signatures of interesting physics, e.g., events with large amounts of deposited energy or unusual combinations of particles, to decide which events in a detector to keep.
%  In the CMS detector, the trigger decides whether to store an event 3.8 $\mu$s after the particle collision has occurred at the center of the detector. 
There are multiple stages of the trigger system, and the first stage, referred to as the level-1 trigger (L1T)~\cite{CMSL1T,ATLASL1T}, processes data at 40\,MHz with custom ASICs or FPGAs.
Over the past years, the LHC has increased its center of mass collision energy and instantaneous luminosity to allow experiments to hunt for increasingly rare signals.
With the extreme uptake of accumulated data, ML methods are being explored for various tasks at the L1T~\cite{CMSP2L1T,CMS-DP-2022-021}.
One such task is \emph{jet tagging}: identifying and classifying collimated showers of particles from the decay and hadronization of quarks and gluons using \emph{jet substructure} information~\cite{Larkoski:2017jix,Kogler:2018hem}.
ML methods show great promise over traditional algorithms in increasing our capability to identify the origins of different jets and discover new physical interactions~\cite{Kasieczka:2019dbj,Sirunyan_2020}.

Within the context of developing a NN for real-time decision making for particle physics applications, the original contributions of this paper are the following:
\begin{itemize}
  \item
        We take advantage of the QONNX format to represent QNNs with arbitrary precision and mixed-precision quantization in order to extend HAWQ for QONNX intermediate representation support.
  \item
        We perform Hessian-aware quantization on a multilayer perceptron (MLP) model used in jet tagging benchmarks, and we study in detail the effects of quantization on each layer for model performance and efficiency.
  \item
        We use hls4ml to present optimized resources and latency for FPGA hardware implementations of NNs trained in HAWQ.
\end{itemize}

The rest of this paper is structured as follows.
In Section~\ref{sec:background-related-work}, we introduce the key steps that comprise the end-to-end codesign workflow for QNNs to be deployed on FPGAs and ASICs, including an overview of quantization and HAWQ.
We present the task and discuss how NNs are evaluated and trained in Section~\ref{sec:setup}.
Preliminary QAT results with homogeneous quantization and Hessian-based quantization are presented in Section~\ref{sec:quantization-aware-training}, and
our extension to HAWQ is presented in Section~\ref{sec:conversion-into-qonnx}.
We then cover the firmware implementation of NNs, specifically the resource usage and estimated latency, in Section~\ref{sec:hardware-generation}.
Finally, a summary is presented in Section~\ref{sec:summary}.

\section{Background and Related Work} 
\label{sec:background-related-work}

In this section, we provide an overview of quantization and HAWQ (in Section~\ref{sec:background}); and then we cover automatic bit width selection (in Section~\ref{sec:background-bit-width}) and firmware generation tools (in Section~\ref{sec:background-firmware-gen}).

\subsection{Quantization}
\label{sec:background}

% ========================== QUANTIZATION & INFERENCE ========================== 
Quantization in NNs refers to reducing the numerical precision used for inputs, weights, and activations.
In \emph{uniform affine quantization}, values are quantized to lower precision integers using a mapping function defined as
\begin{equation}
\label{eq:quantization}
        q = \quantize(r) = \Clip(\Round((r/S) - Z), \alpha, \beta),
\end{equation}
where $r$ is the floating-point input, $S$ is the \emph{scale factor}, and $Z$ is the \textit{zero point}~\cite{gholami2021survey}.
The $\Round$ function is the round-to-nearest operation clipped/clamped at $\alpha$ and $\beta$.
Because all quantization bins are uniformly spaced, this mapping function in Eqn.~\ref{eq:quantization} is referred to as uniform quantization.
Nonuniform quantization methods whose bin sizes are variable are more difficult to implement in hardware~\cite{liu2022nonuniform}.
Real values can be recovered from the quantized values through \emph{dequantization}:
\begin{equation}
\label{eq:dequantization}
        \Tilde{r} = \dequantize(q) = S(q + Z),
\end{equation}
where $\Tilde{r} - r$ is known as the quantization error.
The scale factor divides a given range of real values into $2^b$ bins, with
\begin{equation}
\label{eq:scalefactor}
        S = \frac{\beta-\alpha}{2^b-1},
\end{equation}
where $[\alpha, \beta]$ is the clipping range and $b$ is the bit width.
Choosing the clipping range is referred to as \textit{calibration}.
A simple approach is to use the minimum and maximum of the values, i.e., $\alpha=r_{\min}$, and $\beta=r_{\max}$.
This is an asymmetric quantization scheme because the clipping range is not necessarily symmetric with respect to the input, i.e., it could be that $-\alpha \neq \beta$.
A symmetric quantization approach uses a symmetric clipping range of $-\alpha = \beta$, such as 
$-\alpha =\beta = \max(|r_{\max}|, |r_{\min}|)$, and replaces the \textit{zero point} with $Z = 0$.

The latest publication of the Hessian-aware quantization, HAWQv3 ~\cite{hawqv3}, introduces a completely new computational graph with an automatic bit width selection policy based on it's previous works ~\cite{hawqv1, hawqv2}. 
In HAWQv3, which for simplicity we refer to here simply as HAWQ, quantization follows Eqn.~\ref{eq:quantization} with additional hardware-inspired restrictions.
HAWQ executes its entire computational graph using only integer multiplication, addition, and bit shifting, without any floating-point or integer division operations.
The clipping range is symmetric for weights $\beta=2^{b}-1=-\alpha$, while activations can be either symmetric or asymmetric.
The real-valued scale factors are pre-calculated by analyzing the range of outputs for different batches and fixed at inference time, a process called \emph{static quantization}.
HAWQ avoids floating-point operations and integer divisions by restricting all scale factors to be dyadic numbers (rational numbers of the form $b/2^c$, where $b$ and $c$ are integers).
To illustrate a typical computation, consider a layer with input $h$ and weight tensor $W$.
In HAWQ, $h$ and $W$ are quantized to $S_hq_h$ and $S_Wq_W$, respectively, where $S_h$ and $S_W$ are the real-valued scale factors, and $q_h$ and $q_W$ are the corresponding quantized integer values.
The output result, denoted by $a$, can be computed as
\begin{equation}
\label{eq:matricmult}
        a = (S_W S_h)(q_W \ast q_h),
\end{equation}
where $\ast$ denotes a low-precision integer matrix multiplication (or convolution).
The result is then quantized to $S_a q_a$ for the following layer as
\begin{equation}
\label{eq:scaling}
        q_a = \Int\left(\frac{a}{S_a}\right) = \Int\left(\frac{S_WS_h}{S_a}(q_W\ast q_h)\right),
\end{equation}
where $S_a$ is a precalculated scale factor for the output activation.
This avoids floating point operations and integer divisions by implementing Eqn.~\ref{eq:scaling} with integer multiplication and bit shifting.

% ========================== MIXED-PRECISION QUANTIZATION ========================== 
\subsection{Automatic Bit Width Selection}
\label{sec:background-bit-width}
Many methods have been proposed to measure the sensitivity to quantization or developed automatic schemas for bit settings.
For example, HAQ~\cite{haq} proposed a reinforcement learning (RL) method to determine the quantization policy automatically.
The method involves an RL agent receiving direct latency and energy feedback from hardware simulators.
Ref.~\cite{wu2019mixed} formulated a neural architecture search (NAS) problem with a differentiable NAS (DNAS) to explore the search space efficiently.
Ref.~\cite{Naumov} proposed periodic functions as regularizers, where regularization pushes the weights into discrete points that can be encoded as integers.
One disadvantage of these exploration-based methods is that they are often sensitive to hyperparameters or initialization.
More recently, AutoQkeras~\cite{Coelho2020UltraLL} was proposed as a method to optimize both model area (measured by the number of logical elements in the FPGA design) and accuracy, given a set of resource constraints and accuracy metrics, e.g., energy consumption or bit-size. 
Different from these previous methods, HAWQ~\cite{hawqv1} introduced an automatic way to find the mixed-precision settings based on a second-order sensitivity metric.
In particular, the Hessian (specifically the top Hessian eigenvalue) can be used to measure the sensitivity.
This approach was extended in Ref.~\cite{hawqv2}, where the sensitivity metric is computed using the average of all the Hessian~eigenvalues.

% ========================== FIRMWARE GENERATION ========================== 

\subsection{Firmware Generation Tools}
\label{sec:background-firmware-gen}

Although ML methods have shown promising results on edge devices, fitting these algorithms onto FPGAs is challenging, often very time-consuming, and it requires the expertise of domain experts and engineers.
Several directions aim to solve this issue.
One direction, field-programmable DNN (FP-DNN)~\cite{fpdnn}, is a framework that takes TensorFlow-described deep neural networks (DNNs) as input and automatically generates hardware implementations with register transfer level (RTL) and high-level synthesis (HLS) hybrid templates.
Another direction, fpgaConvNet~\cite{fpgaconvnet}, specifically targets convolutional NNs (CNNs) and is an end-to-end framework for the optimized mapping of CNNs on FPGAs.
Interestingly, fpgaConvNet proposes a multi-objective optimization problem to account for the CNN workload, target device, and metrics of interest.

These and other tools indicate a growing desire to deploy more efficient and larger ML models on edge devices in a faster and more streamlined process.
This desire arises in many scientific and industrial use cases~\cite{banbury2021mlperf,10.3389/fdata.2022.787421}.
Particle physics applications are a particularly strong stress test of such tools.
This is due to the extreme requirements in computational latency and data bandwidth, as well as environmental constraints such as low-power and high-radiation and cryogenic environments.
Furthermore, particle physics practitioners are not necessarily ML experts or hardware experts, and their applications and systems require open-source tools (to the extent possible) and flexible deployment across different FPGA and ASIC platforms.
The hls4ml tool originated from such use cases, and it supports multiple architectures and frameworks, such as Keras~\cite{chollet2015keras}, QKeras~\cite{Coelho2020UltraLL,qkeras}, and PyTorch~\cite{pytorch}.
Currently, it is steadily increasing its scope of supported architectures, frameworks, hardware optimizations, and target devices, with the backing of a growing scientific community.
Another tool, FINN~\cite{blott2018finn,finn} from AMD/Xilinx, aims to solve the problem of bringing NNs (more specifically, QNNs) to FPGAs by using generated high-level synthesis (HLS) code.
Both tools create a streamlined process to deploy DL models as efficiently as possible, without requiring large development effort and time.
The two tools are similar in their goals, hls4ml and FINN, though there are differences in their flows, layer support, and targeted optimizations.
Both of them support QONNX, an open-source exchange format, representing QNNs with arbitrary precision, such that there can be interoperability between the flows.
More generally, we should note that this is ideal for HAWQ, as it can target multiple hardware-generating tools.
In this work, however, we focus only on hls4ml, which has implementations for FPGAs and ASICs and optimizations for a larger range of bit widths.

% ========================== EXPERIMENTAL SETUP ========================== 

\section{Experimental Setup}
\label{sec:setup}

In this section, we describe the benchmark ML task we explore for particle physics applications.
As discussed above, although there are a much broader set of scientific and industrial applications, particle physics applications are a particularly good stress test of our end-to-end workflow.
The concept behind the development of particle physics benchmarks is detailed more in Ref.~\cite{Duarte:2022hdp}, and our jet tagging benchmark is one of the three described there.

% ========================== DATASET ========================== 

\subsection{Dataset}
We consider a jet classification benchmark of high-\pt jets to evaluate performance.
Particle jets are radiation patterns of quarks and gluons produced in high-energy proton-proton collisions at the LHC.
As these jets propagate through detectors like ATLAS or CMS, they leave signals through the various subdetectors, such as the silicon tracker, electromagnetic or hadron calorimeters, or muon detectors.
These signals are then combined using jet reconstruction algorithms.
We use the benchmark presented in Ref.~\cite{fast-inference} consisting of 54 features from simulated particle jets produced in proton-proton collisions.
Of the 54 high-level features, 16 were chosen based on Table 1 of Ref.~\cite{fast-inference}.
The features are a combination of both mass (``dimensionful'') and shape (``dimensionless'') observables.
The dataset~\cite{hls4ml_dataset} is a collection of 870,000 jets and is divided into two sets: a training set of 630,000 jets, and a test set of 240,000 jets.
The dataset underwent preprocessing: all features are standardized by removing the mean and scaling to obtain unit variance.
The task is to discriminate jets as originating from one of five particles: W bosons, Z bosons, light quarks (q), top quarks (t), or gluons (g).
Descriptions of each observable and particle jet can be found in Ref.~\cite{Coleman_2018}.
Additionally, we measure the accuracy given by the number of correctly classified jets divided by the total number of classified jets.

% ========================== MODEL & LOSS DEFINITION ========================== 

\subsection{Model \& Loss Definition}

We implement all models with the architecture presented in Ref.~\cite{fast-inference}, an MLP with three hidden layers of 64, 32, and 32 nodes, respectively.
The baseline model is the floating-point implementation of this MLP, i.e., with no quantization.
All hidden layers use ReLU activations, and the output is a probability vector of the five classes filtered through the softmax activation function.
We aim to minimize the empirical loss function
\begin{align}
    \mathcal{L}_c(\theta) &= \frac{1}{N}\sum_{i=1}^{N}\ell(f_\theta(\textbf{x}_i),\textbf{y}_i)
    =\frac{1}{N}\sum_{i=1}^{N}\ell(\hat{\textbf{y}}_i,\textbf{y}_i),
  \end{align}
where $\ell$ is the categorical cross-entropy loss function and $N$ is the number of training samples.
The model, denoted by $f_\theta$, maps each input $\textbf{x}_i \in \mathbb{R}^{16}$ to a prediction $\hat{\textbf{y}}_i \in [0,1]^{5}$, using parameters $\theta$.
Predictions are then compared with ground truth $\textbf{y}_i$ to minimize the empirical loss.
% \michael{I'm guessing that $x$ and $y$ and $\hat{y}$ should all have $i$ subscripts here?}
We train the NNs with $L_1$ regularization by including an additional penalty term to the loss,
\begin{equation}\label{eq:loss}
  \mathcal{L}(\theta) = \mathcal{L}_c(\theta) + \lambda\sum_{j=1}^{L}\lVert\textbf{W}_j\rVert_1,
\end{equation}
where the added penalty term is the elementwise norms of weight matrices, $\textbf{W}_i$ is the "vectorized" form of weight matrix for the $j^\mathrm{th}$ layer, and $L$ is the number of layers in the model. 
% \michael{I assume that this is the elementwise L1, with the matrix viewed as a vector, not the matrix L1, but clarify.}
The $L_1$ regularization term is scaled by a tunable hyperparameter $\lambda$.
Typically, $L_1$ regularization is used to prevent overfitting, enabling statistical models to generalize better outside the training data.
It is also known to promote sparsity, which is desirable to reduce the number of computations.
Section~\ref{sec:quantization-aware-training} discusses the implications of $L_1$ regularization in QNNs concerning performance and other metrics discussed below.

% ========================== METRICS ========================== 

\subsection{Metrics: Bit Operations \& Sparsity}

% ========================== BIT OPERATIONS ========================== 

Similar to floating-point operations, bit operations (BOPs)~\cite{bops} in QNNs are computed to estimate model complexity and the number of operations per inference.
BOPs have been shown to predict accurately the area of hardware accelerators and, in turn, the power usage in processing elements~\cite{hmc}.
This makes BOPs an easy-to-compute metric that is a useful approximation of the total area of a QNN.
The bit operations of a fully connected layer with $b_a$ bit input activations and $b_W$ bit weights is estimated by:
\begin{equation}\label{eq:bops}
  \mathrm{BOPs} \approx mn((1-f_p)b_ab_W + b_a + b_W + \log_2(n)),
\end{equation}
where $n$ and $m$ are the number of input and output features, and the $(1-f_p)$ term accounts for a fraction of weights pruned (i.e., equal to zero).
From Eqn.~\ref{eq:bops}, the number of BOPs is inversely proportional to the sparsity.
Sparse models are desired, as zero-weight multiplications are optimized out of the firmware implementation by HLS.
This is a highly attractive feature of HLS, and it makes BOPs a noteworthy metric to observe.
We measure the total BOPs of each quantization scheme as well as its relation with accuracy (see Section~\ref{sec:quantization-aware-training}) and hardware usage (see Section~\ref{sec:hardware-generation}).

% ========================== Quantization-aware Training ========================== 

\section{Quantization-Aware Training}
\label{sec:quantization-aware-training}

In this section, we discuss the training procedure for homogeneous and mixed-precision quantization.
We start in Section~\ref{sec:same-bitwidth} with a discussion of single bitwidth quantization, which is also referred to as homogeneous quantization.
Then, in Section~\ref{sec:mixed}, we discuss mixed-precision quantization, including how it can greatly improve classification performance, as well as its downsides.
In particular, in Section~\ref{sec:hessian-aware-quantization}, we cover a method to select automatically the bit width of each layer in a NN using second-order Hessian information, as well as a method obtained by imposing hardware constraints in the bit width selection process.

% ========================== HOMOGENEOUS QUANTIZATION ========================== 

\subsection{Homogeneous Quantization}
\label{sec:same-bitwidth}
Quantizing all layers with the same bit width is simple, but it can cause a significant loss in performance.
In Table~\ref{tab:uniform}, we present the accuracy for different bit settings from INT12 to INT4 with homogeneous quantization using HAWQ.
As expected, we see a significant performance degradation as we quantize below INT8 (and especially below INT6).
To combat this, we employed two regularization techniques during training: $L_1$ regularization and batch normalization (BN)~\cite{batchnorm}.
BN provides a more stable distribution of activations throughout training by normalizing the activations and producing a smoother loss landscape~\cite{batchnorm_opt}.
Although using BN raises performance on all quantization schemes, it fails to recover baseline accuracy for INT6 and INT4 quantization.
Similarly, $L_1$ regularization improves the model somewhat, but it fails to restore performance to its baseline.
Consequently, homogeneously quantizing a model with one bitwidth setting is insufficient for quantization below 8-bit precision.

\begin{table}[h]
  \centering
  \begin{tabular}{llSSSSS} \toprule
    \multicolumn{2}{c}{Precision} & 
    {Baseline [\%]} & 
    {$L_1$ [\%]} & 
    {BN [\%]} & 
    {$L_1$+BN [\%]} 
            \\
    \cmidrule(lr){1-2}
        Weights &  Inputs     &    &  & &   &  \\
    \midrule
    INT12       & INT12   & 76.916          & 72.105      & 77.180   & 76.458         \\
    INT8        & INT8    & 76.605          & 76.448      & 76.899   & 76.879         \\
    INT6        & INT6    & 73.55           & 73.666      & 74.468   & 74.415         \\
    INT4        & INT4    & 62.513          & 63.167      & 63.548   & 63.431         \\
    \midrule
    FP-32       & FP-32   & 76.461          & 76.826      & 76.853   & 76.813         \\
    \bottomrule
  \end{tabular}
  \caption{\label{tab:uniform} Classification performance with homogeneous quantization.
    All weights, activations, and inputs are quantized with the same precision.
    Models are trained with and without $L_1$ regularization and BN.
    At INT8 and above, the accuracy is restored to baseline; but at INT6 and below, the accuracy is worse than baseline.}
\end{table}

\begin{figure*}[ht]
  \centering
  \includegraphics[width=\textwidth]{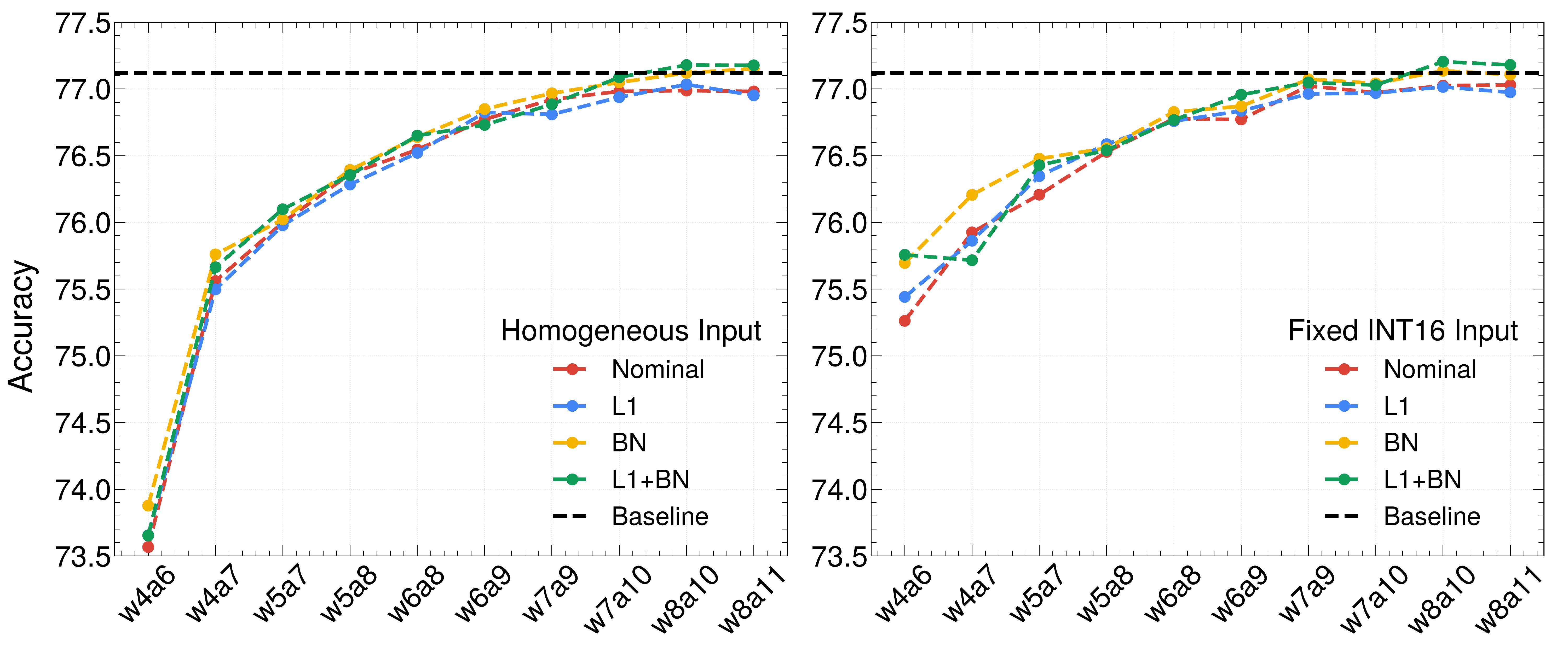}
  \caption{\label{fig:uniform-extend} Model performance using homogeneous quantization.
    The precision of weights is indicated after ``w'' and activations after ``a.''
    Models are trained with $L_1$ regularization and BN.
    We can see: (1) 16-bit input improves the model performance of all bit settings; (2) larger activation bit widths improve accuracy; and (3) $L_1$ and BN (applied alone or together) show no positive impact on performance.}
\end{figure*}

In addition to employing regularization techniques, we can increase the input quantization bit width.
In HAWQ, inputs are quantized before proceeding to the first layer, ensuring all operations are integer only.
A possible failure point is quantization error introduced in the inputs for low bitwidths where key features needed to classify jets may be lost.
We decouple the precision of the inputs from that of the weights and activations and increase it to INT16. 
Fig.~\ref{fig:uniform-extend} shows results for 8-bit weights and below, with different bit widths for the activations.
We find: 
(1) increasing the activation bit width significantly improves the classification performance of INT4 and INT6 weights; 
(2) similar improvements are obtained for INT16 quantized inputs---although this comes at the cost of increased hardware resource usage; and 
(3) $L_1$ and BN (applied alone or together) are insufficient for recovering the accuracy to baseline levels.  
For this study, BN is less desirable, as the batch statistics parameters are implemented with floating-point values, thereby increasing the latency and memory footprint.
One option is to quantize these values or (even more promisingly) apply BN folding.
The idea is to remove BN by using its parameters to update the fully connected (or convolution) layer's weights and biases for inference efficiency.
However, after we explored BN folding using the procedure outline in Ref. ~\cite{hawqv3}, we found little to no effect on model performance.
As previously mentioned, $L_1$ produces sparse matrices, decreasing the number of bit operations needed in hardware.
Henceforth, in later sections, we continue to use $L_1$ during training for mixed-precision quantization.
Fig. \ref{fig:uniform-extend} suggests model performance can greatly benefit from more fine-grained quantization settings.
However, manually adjusting all these quantization settings can be time-consuming and suboptimal.
An optimized bit-setting scheme is needed to simultaneously minimize the loss and hardware usage.
In the next subsection, we explore mixed-precision quantization.
We fix the input bit width to INT16.
This could be further optimized, but this choice makes direct comparison with other work easier~\cite{Duarte:2022hdp,fast-inference,Coelho2020UltraLL,Hawks:2021ruw}.

% ========================== MIXED-PRECISION QUANTIZATION ========================== 

\subsection{Mixed-Precision Quantization} \label{sec:mixed}
\subsubsection{Brute-force Search}

Mixed-precision quantization aims to improve performance by keeping certain layers at a higher precision than others.
The basic problem with going beyond homogeneous quantization is that---when implemented naively---the search space for determining the bit setting is exponential to the number of layers.
Our model architecture's MLP search space is significantly smaller than deep CNNs such as ResNet-50~\cite{resnet50} because our MLP only has 3 hidden layers.
However, assuming we have 5-bit width options, finding the mixed-precision setting for our MLP classifier, with 4 fully-connected layer weights and activations, has a search space of $((2)(4))^5 = 32,768$ combinations.
It is impractical, especially for applications that need frequently retrained models or that need DNNs, to search this space exhaustively.
Several methods have been proposed to address this problem of manually searching for the optimal bit configuration~\cite{shen2020q, cai2020zeroq, haq, wu2019mixed, hawqv2}.
We use Ref.~\cite{hawqv2}, which is based on the Hessian information, and we observe the relative position of Hessian-based solutions within the \textit{brute-force search} space.

% ========================== HESSIAN AWARE QUANTIZATION ========================== 

\subsubsection{Hessian-Aware Quantization}
\label{sec:hessian-aware-quantization}

\begin{figure}[h]
  \centering
  \includegraphics[width=.4\textwidth]{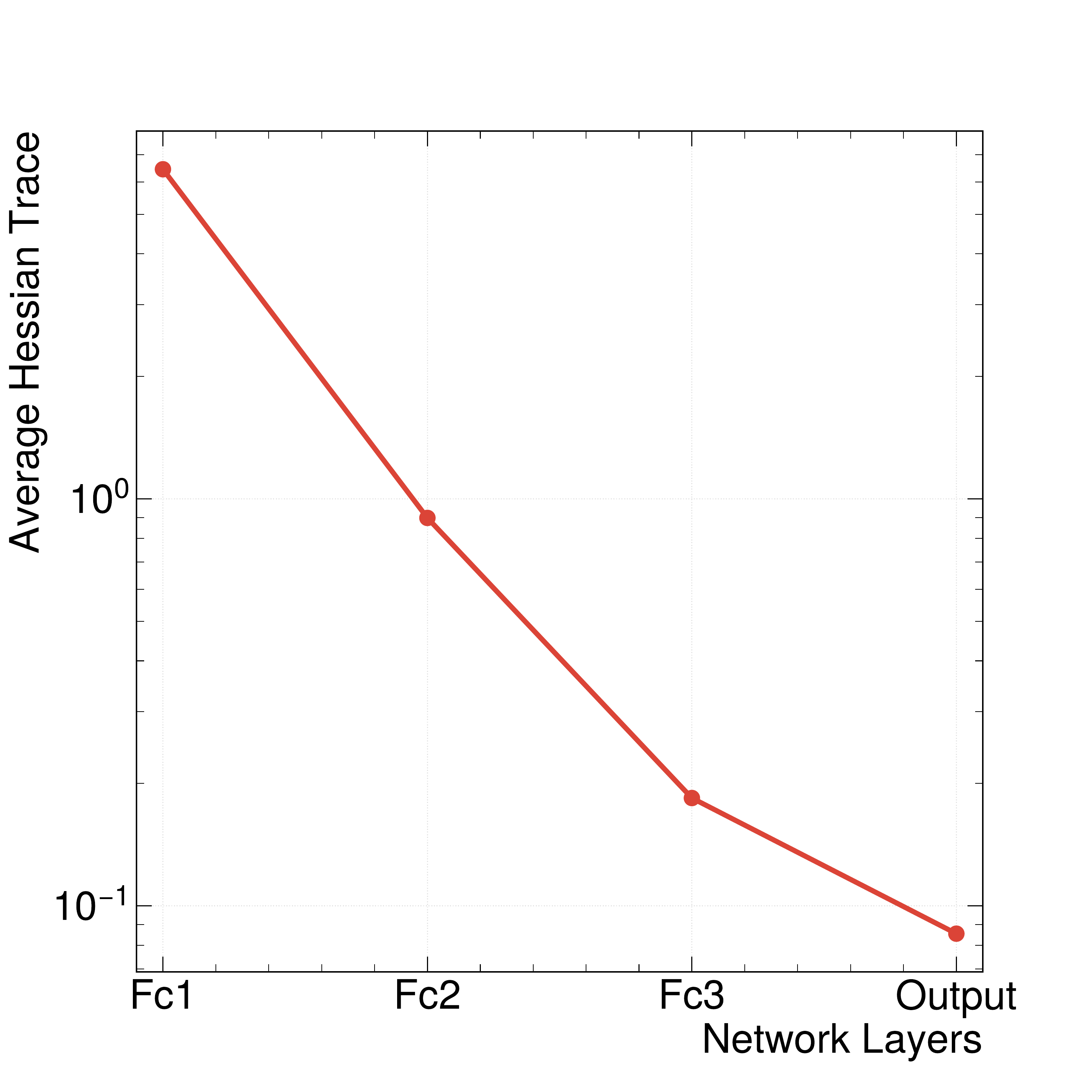}
  \caption{
  \label{fig:trace}
  Average Hessian trace of each fully-connected (Fc) layer in the MLP. The Hessian is used as a sensitivity metric to quantization, where layers are ranked based on their trace.
  The first two layers are significantly larger than the others, signifying they are more prone to error at lower bit widths. 
  The average Hessian traces are used to assign each layer a bit setting, i.e., layers with higher traces are assigned larger precision.
  }
\end{figure}

As discussed in the Sec.  ~\ref{sec:same-bitwidth}, performance greatly benefited from higher precision in activations suggesting certain layers are more sensitive to quantization than others.
We use the work first proposed in HAWQv2~\cite{hawqv2} to determine the relative sensitivity of each layer for the baseline 32-bit floating point implementation of the model.
The sensitivity metric is computed using the Hutchinson algorithm,

\begin{equation} 
\label{eq:trace}
  \mathrm{Tr}(H) \approx \frac{1}{k} \sum_{i=1}^{k} z_{i}^{\intercal} H z_{i} = \mathrm{Tr}_\mathrm{Est}(H),
\end{equation}
where $H\in \mathbb{R}^{d\times d}$ is the Hessian matrix of second-order partial derivatives of the loss function with respect to all $d$ model parameters, $z\in \mathbb{R}^d$ is a random vector whose component is i.i.d. sampled Rademacher distribution, and $k$ is the number of Hutchinson steps used for trace estimation.
Fig.~\ref{fig:trace} shows the average Hessian trace (our sensitivity metric) of each layer in the baseline model, with logarithmic scaling.
The first two layers are the most sensitive, with the first layer more sensitive than the second by a factor of 7.
Thus, the first two layers in the network must have a larger bit width setting,  while the last two layers can be quantized more aggressively.
While the Hessian traces provides a sensitivity metric, this does not directly translate to a bit configuration.
Instead, Ref.~\cite{hawqv2} assigns the bit width of each layer $i$ by checking the corresponding $\Omega$ term, defined as:

\begin{equation} 
\label{eq:trace-omega}
  \Omega = \sum_{i=1}^{L} \Omega_{i} = \sum_{i=1}^{L}\overline{\mathrm{Tr}}(H_i)  \lVert Q(W_i)-W_i\rVert_{2}^{2},
\end{equation}
where $Q$ is the quantization function, $\lVert Q(W_i)-W_i\rVert_{2}^{2}$ is the squared $L_2$ norm of the quantization perturbation, and $\overline{\mathrm{Tr}}$ is the average Hessian trace.
We apply the same technique as Ref.~\cite{hawqv2}, where the amount of second-order perturbation, $\Omega$, is calculated for a given set of quantization schemes, and the minimal $\Omega$ is chosen. 
This procedure is fully automated without any manual intervention.

We follow the procedure outlined in HAWQ~\cite{hawqv3} to constrain Eqn.~\ref{eq:trace-omega} by the total BOPs.
We formulate an integer linear programming (ILP) optimization problem, where the objective is to minimize $\Omega_i$ while satisfying the constraints.
We set up an ILP problem to automatically determine the bit settings of our classifier for various BOPs limits, and we compare these solutions with the brute force and homogeneous quantization methods.

% ========================== QUANTIZATION RESULTS ========================== 

\subsubsection{QAT Results}
\label{sec:quantization-results}

\begin{figure*}[t]
  \centering
  \includegraphics[width=\textwidth]{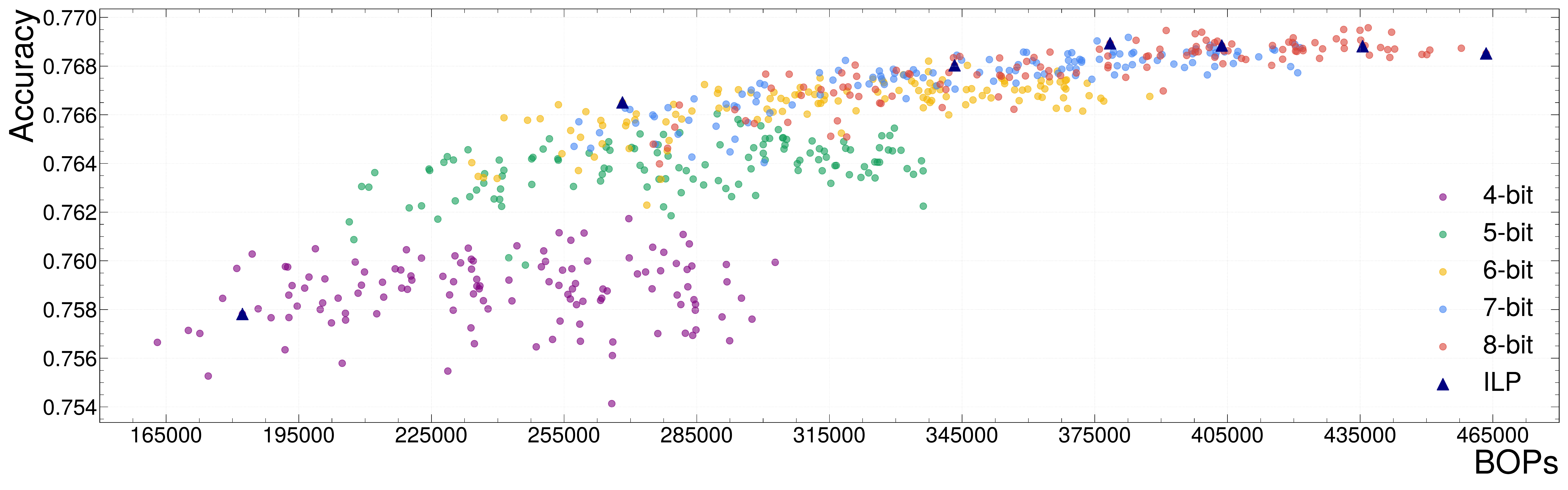}
  \caption{\label{fig:brute-force}Brute-force search quantization using weight bit widths $b_w=\{4,5,6,7,8\}$.
    Each data point is color-coded based on the bit width of the first fully-connected layer.
    It's importance in quantization coincides with the observed clusters, with higher performing models using larger bit widths.
    Solutions based on ILP are presented.
    All ILP solutions make trade-offs based on the quantization error and bit width, and typically are among the lowest BOPs in their respective cluster.
    }
\end{figure*}

With the information provided in Fig.~\ref{fig:uniform-extend}, we apply all possible bit settings based on the initial implementation in homogeneous quantization.
We explore the weight bit width $b_W=\{4,5,6,7,8\}$, and we set the activation bit width $b_a=b_W+3$ to prevent saturation and further reduce the search space.
All models are trained for 100 epochs, with $L_1$ regularization, and all models use quantized inputs with INT16.
Fig.~\ref{fig:brute-force} presents the model accuracy against BOPs for all combinations of weight bits $b_W$.
Data points are color-coded based on the bit precision of the first layer.
Several data points indicate a complete or nearly complete recovery to baseline accuracy (76.853\%).
The majority of points can be clustered based on the bit width of the first layer, since the model accuracy generally increases as the first layer's bit width increases.
We can also see in Fig.~\ref{fig:brute-force} that the bit width of the first fully-connected layer greatly impacts the final model performance.
Among the top 100 best-performing models, 66 had the first dense layer as INT8, and 33 had INT7 weights.
This coincides with the average Hessian traces shown in Fig.~\ref{fig:trace}, showing the first layer is the most sensitive layer to quantization, by a factor of 7$\times$, compared to the second most sensitive layer.
Among the top models, we observed the frequency of 7-bit and 8-bit in later layers decrease significantly.
The bit width of the later layers has fewer effects on the classification than the first two layers.

The ILP solutions to Eqn.~\ref{eq:trace-omega} are also shown.
The solutions are obtained with respect to 7 different BOPs constraints, from 250\,k to 550\,k in steps of 50\,k.
As expected, as the BOPs constraint increases, the selected precision of the first two layers increases.
Hence, we begin to see more ILP solutions closer to the 8-bit cluster.
The ILP solutions also tend to be positioned towards the lower end of BOPs in their local cluster.
With brute-force search quantization and the ILP solutions shown side by side, the advantages of using the Hessian information become clearer.
While an optimal solution is not guaranteed, the Hessian provides a stable and reliable solution to mixed-precision quantization.
This is ideal for deep learning models that need to be quantized to meet the resource constraints and inference times of the LHC 40\,MHz collision rate.

% ========================== CONVERSION INTO QONNX ========================== 

\section{Conversion into QONNX} 
\label{sec:conversion-into-qonnx}

\subsection{Intermediate Representations}
To increase interoperability and hardware accessibility, the Open Neural Network Exchange (ONNX) format was established to set open standards for describing the computational graph of ML algorithms~\cite{bai2019}. 
ONNX defines a common and wide set of operators enabling developers and researchers greater freedom and choice between frameworks, tools, compilers, and hardware accelerators. 
Currently, ONNX offers some support for quantized operators, including \texttt{QuantLinear}, \texttt{QLinearConv}, and \texttt{QLinearMatMul}.
However, ONNX falls short in representing arbitrary precision and ultra-low quantization, below 8-bit precision. 
To overcome these issues, recent work~\cite{Pappalardo:2022nxk} introduced quantize-clip-dequantize (QCDQ) using existing ONNX operators and a novel extension with new operators, called QONNX, to represent QNNs.
QONNX introduces three new custom operators: \texttt{Quant}, \texttt{Bipolar}, and \texttt{Trunc}. 
The custom operators enable uniform quantization and abstract finer details, making the intermediate representation graph flexible and at a higher level of abstraction than QCDQ.

For these reasons, we represent HAWQ NNs in the QONNX format, leveraging HAWQ's ultra-low precision and QONNX's abstraction to target two FPGA synthesizing tools, hls4ml and FINN~\cite{blott2018finn, finn}.\footnote{The main focus in exporting QNNs is the QONNX intermediate format. 
However, the QONNX software toolkit enables conversion to QCDQ format.
This allows HAWQ to target hls4ml and FINN, and indirectly all other ONNX inference accelerators and frameworks.}
We also include the ONNX format in our model exporter for representing QNNs.
In the next subsections, we describe the setup, export procedure, and validation steps to represent HAWQ NNs in the QONNX and QCDQ intermediate representations. 

% ========================== MODEL TRANSLATION ========================== 

\subsection{Model Translation} 
\label{sec:translation}

In PyTorch, exporting to ONNX works via tracing. 
This is the process of capturing all the operations invoked during the forward pass on some input.
PyTorch provides the means for tracing through the \texttt{torch.jit} API. 
Tracing a model will return an executable that is optimized using the PyTorch just-in-time compiler.
The executable contains the structure of the model and original parameters. 
Tracing will not record any control flow like if-statements and loops. 
The returned executable will always run the same traced graph on any input, which may not be ideal for functions or modules that are expected to run different sets of operations depending on the input and model state.
The executable is then used to build the ONNX graph by translating operations and parameters within the executable to standard ONNX operators.
In general, all PyTorch models are translated to ONNX using this process, and we extend this existing system to build support for QONNX operators in HAWQ.

The layers in HAWQ and operators in QONNX both require extra steps to support tracing and export.
For each quantized layer in HAWQ, we implement a corresponding ``export'' layer.
These dedicated export layers implement the forward pass and specify the equivalent QONNX operators based on the original layer parameters.
This is accomplished by registering \emph{symbolic functions} via \texttt{torch.onnx.register\_custom\_op\_symbolic}. 
These symbolic functions decompose HAWQ layer operations into a series of QONNX nodes.
Because we are using custom QONNX nodes, we also must register them via the \texttt{torch.onnx} API. 
Together, these preliminary steps define the HAWQ-to-QONNX translation.
During the export process, the exporter looks for a registered symbolic function for each visited operator.
If a given model contains quantized HAWQ or standard PyTorch layers, it can be traced and finally translated to standard ONNX and QONNX operators. 
Because tracing records computations, the input can be random as long as the dimensions and data type are correct.
The model exported with ONNX and QONNX operators is shown in Fig.~\ref{fig:mlp}.
With these additions, our exporter can perform the following:
\begin{enumerate}
\item export models containing HAWQ layers to QONNX, with custom operators to handle a wide range of bit widths while keeping the graph at a higher level of abstraction; and
\item export models containing HAWQ layers to standard ONNX with INT8 and UINT8 restrictions. 
\end{enumerate}

% ========================== POST-EXPORT ========================== 

\subsection{Post-Export}
\subsubsection{Optimization}

\begin{figure}
     \centering
     \begin{subfigure}[t]{0.3\textwidth}
         \centering
         \includegraphics[width=0.8\textwidth]{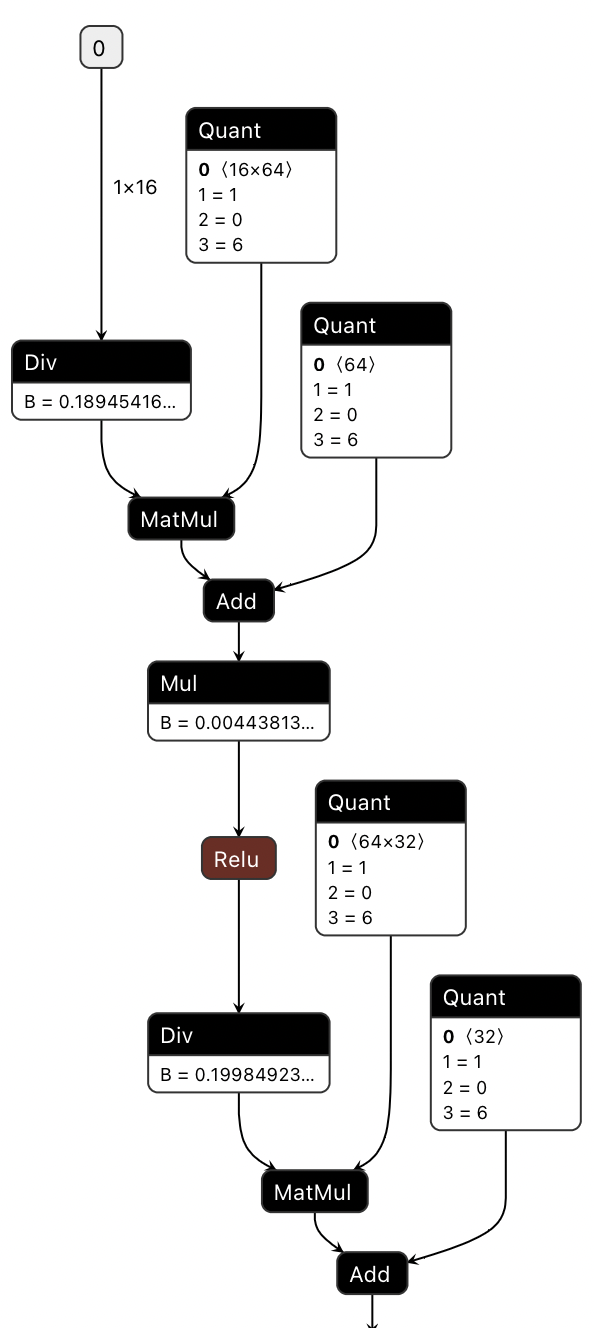}
         \caption{}
         \label{fig:mlp}
     \end{subfigure}
     \hfill
     \begin{subfigure}[t]{0.3\textwidth}
         \centering
         \includegraphics[width=0.8\textwidth]{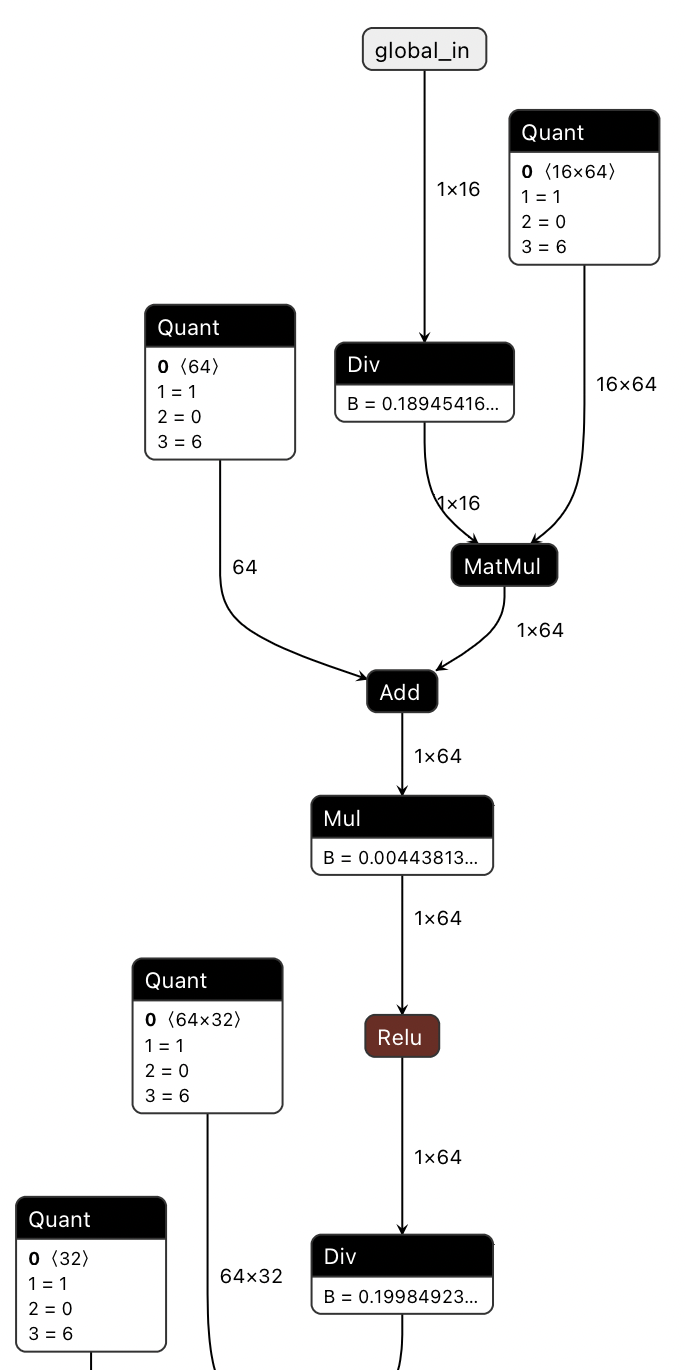}
         \caption{}
         \label{fig:mlp-clean}
     \end{subfigure}
     \hfill
     \begin{subfigure}[t]{0.3\textwidth}
         \centering
         \includegraphics[width=0.8\textwidth]{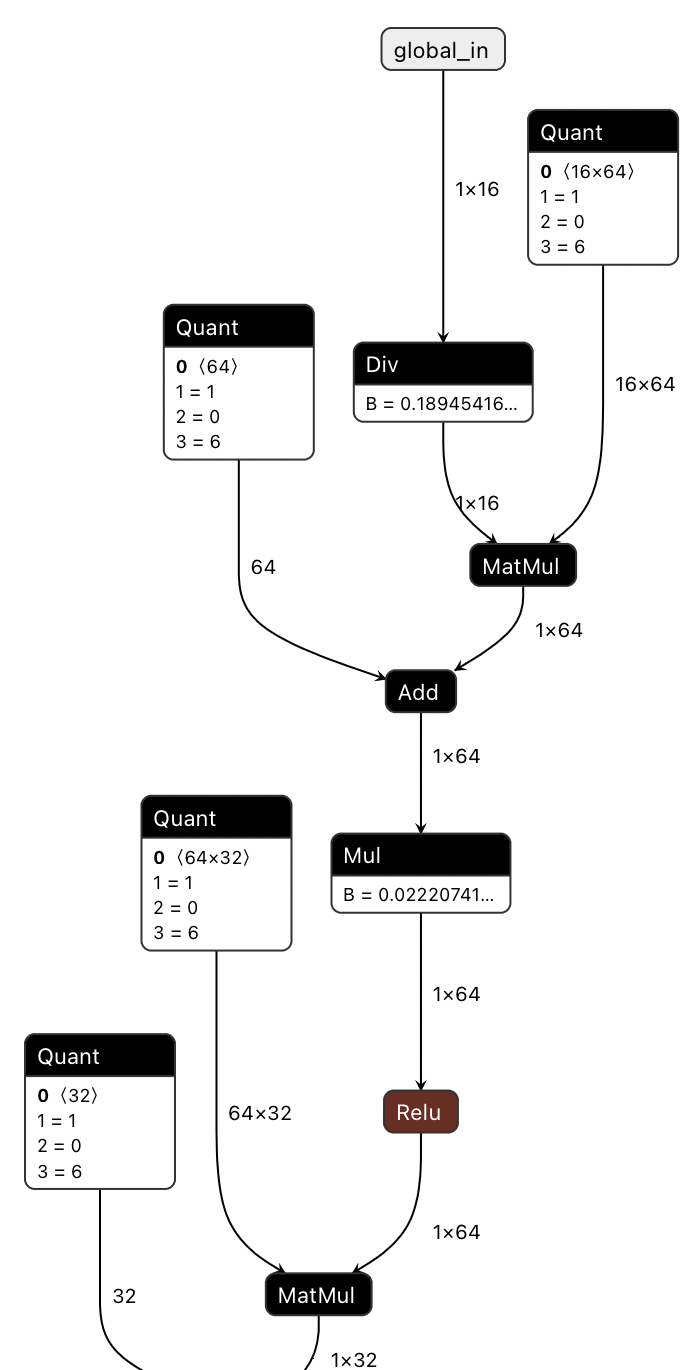}
         \caption{}
         \label{fig:mlp-opt}
     \end{subfigure}
        \caption{The QONNX graph in its three stages after exporting. 
        (a) The first layers of the model including the quantized fully-connected layer before any optimizations. 
        (b) The first layers of the model after post-clean-up operations: constant folding, shape inference, tensor, and node renaming. 
        (c) The final optimization step: node merging across ReLU activations.
        All QNNs implemented in HAWQ can be exported to an QONNX or ONNX intermediate representation and undergo transformations as described in each stage.}
        \label{fig:qonnx graphs}
\end{figure}

In order to create firmware using hls4ml or FINN, the QONNX graph is expected to be normalized, i.e., to undergo several optimization steps.
The QONNX software utilities~\cite{Pappalardo:2022nxk} provide these transformations, as shown in Fig.~\ref{fig:mlp-clean}, where shape inference and constant folding are applied to the graph. 
Fig.~\ref{fig:mlp-opt} shows the last optimization step; we merge scaling factors across ReLU activation functions. 
For reasons related to the underlying implementation of HAWQ, there are two scaling operations before and after specific layers. 
For a detailed explanation of these scaling factors, see Section~\ref{sec:background}.
To reduce the number of operations needed in firmware we combine the scaling factors in cases where the ReLU function is used. 
This cannot always be done, and it is dependent on the activation function used. 

% ========================== GRAPH EVALUATION ========================== 

\subsubsection{Graph Evaluation}

After exporting, we evaluate the model using the QONNX software package~\cite{Pappalardo:2022nxk}, confirming a successful translation of our model from HAWQ to QONNX. 
While the main focus has been MLPs, exporting is not limited to this one architecture. 
All HAWQ layers now support QONNX export via the implemented symbolic functions.
Moreover, with the QONNX software package, it is easy to transform, optimize, evaluate, and validate the exported HAWQ models.

% ========================== HARDWARE GENERATION ==========================

\section{Hardware Generation}
\label{sec:hardware-generation}

In this section, we explain where HAWQ fits within the hls4ml hardware generation workflow.
The total resources used, BOPs, and classification performance for different bit width configurations are shown and discussed.

% ========================== HLS4ML INGESTION ========================== 

\subsection{hls4ml Ingestion}

The hls4ml workflow automatically performs the translation of the architecture, weights, and biases of NNs, layer by layer, into code that can be synthesized to RTL with HLS tools.
The first part of this workflow entails training a NN for a task as usual with PyTorch, Keras, QKeras, or HAWQ.
For HAWQ, a QONNX graph must be exported from the model, but this step can (optionally) be performed for all the frameworks (and, eventually, this will be the preferred flow).
Next, hls4ml translates the QONNX graph into an HLS project that can subsequently be synthesized and implemented on an FPGA or ASIC in the final step of the workflow.

All results presented are synthesized for a Xilinx Kintex Ultrascale FPGA with part number \texttt{xcu250-figd2104-2L-e}.
We report the usage of different resources: digital signal processor units (DSPs), flip-flops (FFs), and look-up tables (LUTs).
We do not report the block RAM (BRAM), a dense memory resource, usage because its only use in the design is to store precomputed outputs for the softmax activation, whose numerical precision is the same for all quantization schemes.
Only the ``bare'' firmware design needed to implement the NN is built with RTL synthesis using Vivado 2020.1.
All NNs are maximally parallelized.
In hls4ml, parallelization is configured with a ``reuse factor'' that sets the number of times a multiplier is used to compute the layer's output.
A fully parallel design corresponds to a reuse factor of one.
All resource usage metrics are based on this ``bare'' implementation after RTL synthesis, and  
all designs use a clock frequency of 200\,MHz.

% ========================== SYNTHESIS RESULTS ========================== 

\subsection{Synthesis Results}

\begin{figure*}[ht]
  \centering
  \includegraphics[width=\textwidth]{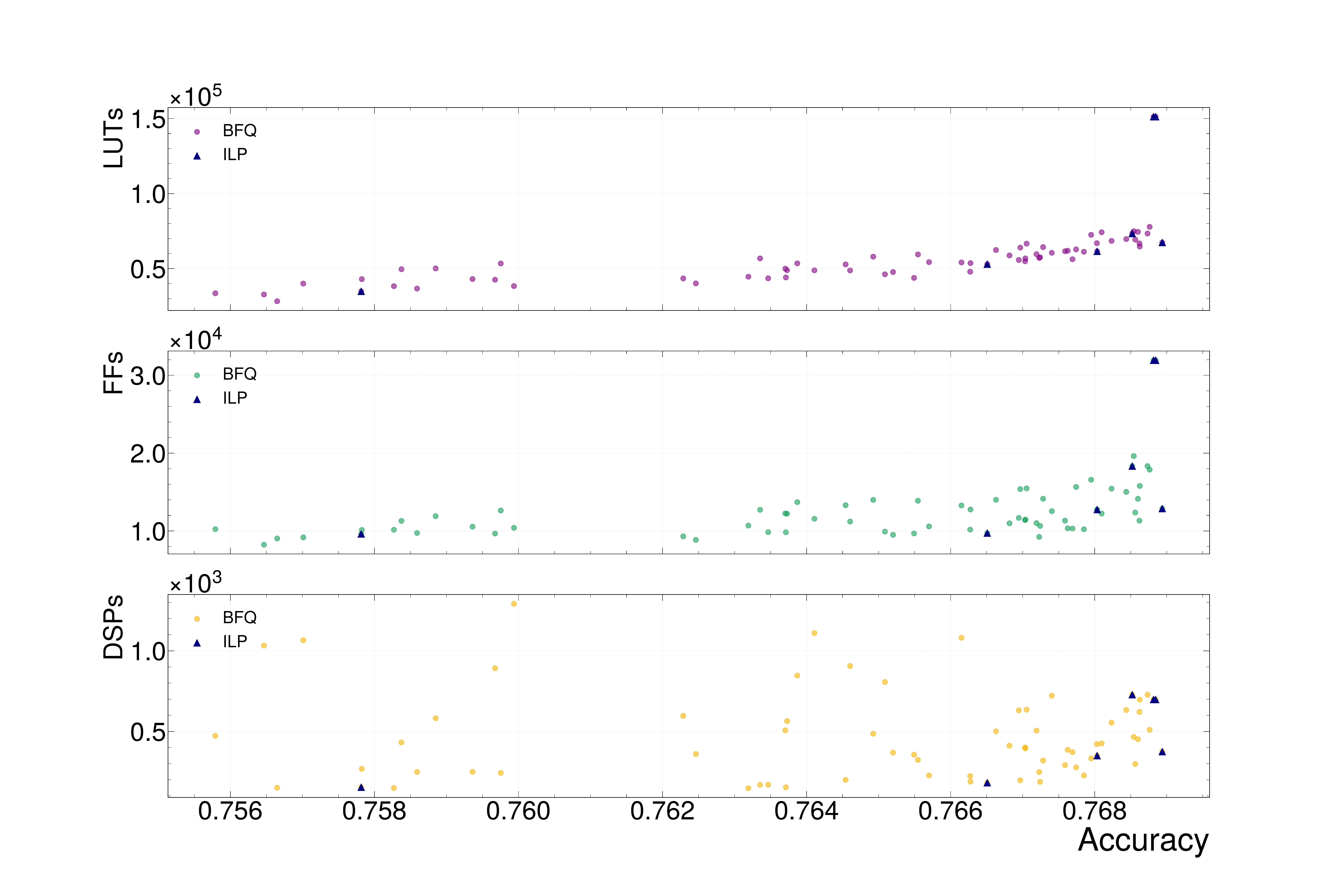}
  \caption{\label{fig:resource-acc}Resource usage for a subset of brute-force quantization (BFQ) using weight bit widths $b_W=\{4,5,6,7,8\}$.
    LUT, FF, and DSP usage versus accuracy are shown, with higher-performing quantization schemes among the highest resource users. 
    All solutions to the ILP problem from BOPs constraint are presented.
    Extra logical elements are needed to maintain accuracy while considerable reduction in all metrics can be achieved with 1-2\% drop in accuracy.}   
\end{figure*}

\begin{figure*}[ht]
  \centering
  \includegraphics[width=\textwidth]{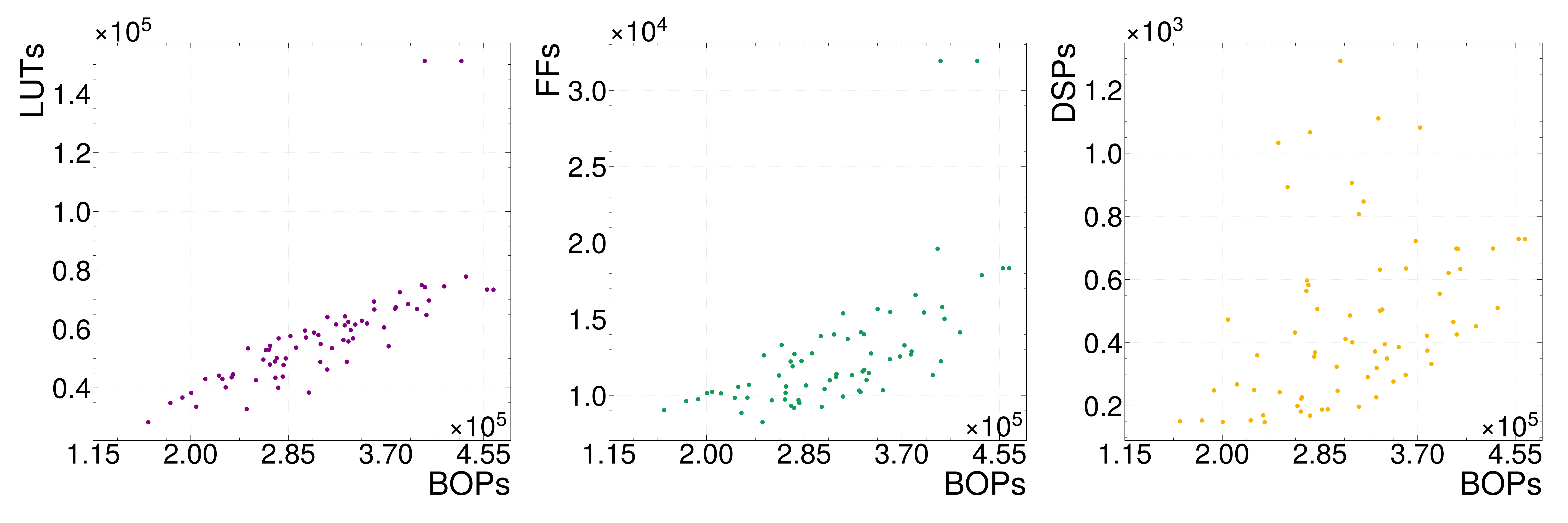}
  \caption{\label{fig:bops-vs-resources}Resource usage for a subset of QNNs in a brute-force attempt to an optimal mixed-precision quantization scheme.
  LUT, FF, and DSP usage versus BOPs are shown, with LUTs having the most linear relationship to BOPs. 
  This relationship weakens with larger bit widths as DSPs can implement MAC operations more efficiently. 
  In all designs, FFs are the only type of memory utilized in fully-connected layers and the total used can drastically vary for neighboring BOPs, implying a weaker relationship between the two.}
\end{figure*}

Fig.~\ref{fig:resource-acc} shows the resource usage compared with the accuracy of the implemented designs.
Quantization bit width settings were chosen at random.
Higher-performing models use more resources.
This is expected, as the top 100 performing models use larger bit widths for the first layer, which is the largest layer in the model.
As such, we expect to see more resources as accuracy increases.
LUTs have the most linear relationship with accuracy, while FFs and DSPs also increase with accuracy.
The relationship between BOPs and resources, presented in Fig.~\ref{fig:bops-vs-resources}, also shows a linear relationship between LUTs and BOPs, which scale with the bit width and weight matrix dimensions.
The number of LUTs used is dependent on the bit width because, at low bit widths, addition and multiplication are implemented with LUTs.
However, DSPs are used at larger bit widths because they become much more efficient.
DSPs offer custom datapaths, efficiently implementing a series of arithmetic operations, including multiplication, addition, multiply-accumulate (MAC), and work-level logical operations.
DSP datapaths are less flexible than programmable logic, but they are more efficient at multiplying and MAC operations.
This is shown in Fig.~\ref{fig:bops-vs-resources} as DSPs usage increases, dramatically at points, with larger BOPs.
Switching from LUTs to DSPs depends on the target device and Vivado HLS internal biases toward DSPs for certain bit widths. 
The shift towards DSPs occurs with 11 or wider bits in Vivado 2020.1, with multiplications lower than this limit implemented using LUTs. 
The result of these operations is stored in FFs, displaying a steady increase with fewer variations than that seen in DSPs.
The number of FFs up to 250\,k BOPs rise at a constant pace with deviations beginning to appear thereafter. The inconsistencies for the number of FFs for neighboring BOPs suggests there is a weaker correlation between the two. The deviations comes from the precision needed for intermediate accumulations and the total FFs needed will vary network to network.

The baseline (BL) model is synthesized after adjusting the weights without any fine-tuning.
In hls4ml, parameters and computations are performed using fixed-point arithmetic, and each layer in the model can be quantized after training by specifying a reduced precision.
Fixed-point data types model the data as an integer and fraction bits with the format \texttt{ap\_fixed<W,I>}.
The BL model uses \texttt{ap\_fixed<16,6>} for all parameters and computations and is fully unrolled, i.e., maximally parallelized, as in previous results.
We compare the BL logical synthesis results with the homogeneous and a Hessian-aware quantization model.
From Table~\ref{tab:uniform}, homogeneous quantization begins to decline below INT8, and we use this quantization scheme to compare to BL.
Of the multiple Hessian-aware solutions, we choose the solution given by the lowest BOPs constraint, i.e., the quantization scheme is 4, 4, 5, and 4 bits for the first, second, third, and final output layers, respectively.
Table~\ref{tab:final-results} shows the synthesis results for three models: BL, INT8 homogeneous quantization, and the Hessian-aware solution.
With INT8 homogeneous quantization, there is a significant reduction in DSPs compared to the BL model, which is further reduced with mixed Hessian-aware quantization.
We expect that as bit width increases, more MAC operations will be implemented in DSPs, which offer a much more efficient implementation than LUTs and FFs.
Interestingly, there's only a minor decrease of FFs with INT8 from BL, compared to the other resources, but this is mostly attributed to INT16 inputs. 
Simply put, larger inputs require more FFs to store and accumulate computations, but they're utilization drastically decrease with lower bit widths.
The MLP with a Hessian-aware quantization scheme uses 42.2\% fewer LUTs, 36.3\% fewer FFs, and 95.7\% fewer DSPs, compared to BL.
As precision is reduced, the number of LUTs needed to compute outputs decreases.
Most computations with lower precision can be implemented with LUTs; hence they have the strongest correlation with BOPs.
However, this observed relationship weakens as bit width increases and DSPs are used instead.
The sudden uptick in LUTs and FFs are outliers that originate from the softmax activation. 
As previously mentioned, the softmax activation stores precomputed outputs and the sudden surge comes from lookup tables created to store all values with large bit widths. 
Table ~\ref{tab:final-results} also includes the automatic mixed-precision solution, QB, from AutoQkeras \cite{Coelho2020UltraLL}, a QNN optimized by minimizing the model size in terms of bits. 
The AutoQkeras solution for jet-tagging, denoted as QB in Table ~\ref{tab:final-results}, drives down all resource metrics by a substantial amount by employing below 4-bit quantization. 
The advantages are also seen in latency while accuracy only drops by a tolerable ~4\%. 
In this study binary and ternary quantization was not explored as in AutoQkeras, but the total gains by leveraging mixed-precision are clearly shown.  

The latency for these models, as estimated by Vivado HLS, is also shown in Table~\ref{tab:final-results}.
Latency estimates are based on the specified clock, the loop transformations' analysis, and the design's parallelization.
Pipelining and data flow choices can heavily change the actual throughput.
However, the latency for the quantized models is about 30\,ns longer than for BL.
This can primarily be attributed to the additional scaling operations of the intermediate accumulations needed for lower precision quantities.
While the additional computation creates an additional latency, the needed resources of these scaling layers are rather modest, 
approximately 1--3\%   
relative to the rest of the design.
So there is a latency-resource trade-off for the lower-precision computations.
However, for the task at hand, the large reduction in resources is worth the increase in latency.
The softmax activation is the other significant contributor to latency, with an estimated 10 ns runtime for all three quantized models presented in Table \ref{tab:final-results}.
As stated above, BRAMs are used for storing the precomputed outputs and the latency mainly arises from reading memory. 
Removing the softmax activation function from the implemented design is usually possible, especially if only the top-$k$ classes are needed for further computation.

\begin{table}
  \centering
  \begin{tabular}{ccccccccc}
    \toprule
    Model                         &
    Acc. [\%]                     &
    Latency [ns]                  &
    \multicolumn{3}{c}{Resources} &
    Sparsity [\%]                 &
    BOPs                                                                                    \\
    \cmidrule(lr){4-6}
                                  &       &    & LUTs   & FFs    & DSPs  &    &           & \\
    \midrule
    Basline                       & 76.85 & 65 & 60,272 & 15,116 & 3,602 & 0  & 4,652,832 & \\
    INT8                          & 76.45 & 95 & 54,888 & 14,210 & 671   & 30 & 281,277   & \\
    Hessian                       & 75.78 & 90 & 34,842 & 9,622  & 154   & 33 & 182,260   & \\
    QB                         & 72.79 & 60 & 16,144 & 4,172  & 5   & 23 & 122,680   & \\
    \bottomrule
  \end{tabular}
  \caption{\label{tab:final-results} 
  Resource usage of the jet-tagging model with different quantization schemes is reported.
    Baseline (no quantization) achieves the highest accuracy with the most resources.
    The same bit width quantization with INT8 reduces DSP and LUT usage, and Hessian-aware quantization significantly reduces all resource metrics. The mixed-precision model, QB, minimizing the total bits from AutoQkeras is shown. Both automatic solutions remove a considerable number of DSPs and LUTs needed for computations, and FFs to store intermediate accumulations. The Hessian is based on the ILP solution from the lowest BOPs constraint.}
\end{table}

\section{Summary}
\label{sec:summary}

The possible applications of HAWQ on edge devices and its automatic bit-setting procedure make it a convincing candidate for physics research.
In this paper, we contributed to the HAWQ library by introducing an extension to convert NNs to ONNX and QONNX intermediate formats.
Bridging HAWQ with firmware synthesis tools that ingest these formats make it easier to deploy NNs to edge devices, such as FPGAs or ASICs, opening many potential use cases in science.
As an initial case study, we employed a NN to classify jets using a challenging benchmark commonly used for QNNs in jet tagging.
We show that the Hessian-aware solution to a mixed precision quantization scheme provides a reliable solution.
We then used our new exporter in HAWQ to translate multiple MLPs optimized with various bit settings to their QONNX IR.
Models were successfully translated from HAWQ to a firmware implementation, and we've observed the resource usage compared to the total BOPs and accuracy.
Furthermore, we compared the resource utilization of multiple different bit settings with the automatic bit selection process in Ref.~\cite{hawqv2}; and
we compared the Hessian-aware model with a homogeneous bit configuration and baseline.
The Hessian-aware solution significantly reduced all resource metrics (LUTs, FFs, and DSPs), with the most significant improvements in DSPs and LUTs, using 95.7\% and 42.2\% fewer DSPs and LUTs compared to baseline, respectively.
Although the current study is limited to MLPs, all NN architectures can first be exported to an ONNX or QONNX intermediate representation graph, and then be applied to whichever tools supports the format.

\section*{Acknowledgmenets}

JC, NT, 
AG, MWM, 
and JD are supported by the U.S. Department of Energy (DOE), Office of Science, Office of Advanced Scientific Computing Research under the ``Real‐time Data Reduction Codesign at the Extreme Edge for Science'' Project (DE-FOA-0002501).
JM is supported by Fermi Research Alliance, LLC under Contract No. DE-AC02-07CH11359 with the DOE, Office of Science, Office of High Energy Physics.
JD is also supported by the DOE, Office of Science, Office of High Energy Physics Early Career Research program under Grant No. DE-SC0021187, and the U.S. National Science Foundation (NSF) Harnessing the Data Revolution (HDR) Institute for Accelerating AI Algorithms for Data Driven Discovery (A3D3) under Cooperative Agreement No. OAC-2117997. NT is also supported by the DOE Early Career Research program under Award No. DE-0000247070,

% \begin{acks}
% \end{acks}

\bibliographystyle{ACM-Reference-Format}
\bibliography{bibliography.bib}

\appendix

\end{document}